\documentclass[DTMColor]{ROB-New}
\ifdefined\pdfminorversion\pdfminorversion=7\fi

\usepackage{array}
\usepackage{bm}
\usepackage{mathrsfs}
\usepackage{etoolbox}
\setcitestyle{numbers,square,comma}
\usepackage{algorithm}
\usepackage{algpseudocode}
\usepackage{colortbl}

\makeatletter

\def\ps@titlepage{%
  \def\@oddhead{}%
  \def\@evenhead{}%
  \def\@oddfoot{}%
  \def\@evenfoot{}%
}

\def\ps@headings{%
  \def\@oddhead{\hfill\thepage}%
  \def\@evenhead{\thepage\hfill}%
  \def\@oddfoot{}%
  \def\@evenfoot{}%
}

\def\printhistory{}

\patchcmd{\@maketitle}
  {\includegraphics{CrossRef-logo}}{}{}{}
\patchcmd{\@maketitle}
  {\href{https://crossmark.crossref.org/dialog?doi=\@logourl}}{}{}{}

\makeatother

\begin{document}

\authormark{DFIO}
\articletype{}
\title[Differentiable Filtering for Inertial Odometry]{Differentiable Filtering for Inertial Odometry}

\authormark{Y. Gu and T. Zou}
\author[1]{Yaru Gu}
\author[1]{Ting Zou\hyperlink{corr}{*}}
\address[1]{Department of Mechanical and Mechatronics Engineering,
   Memorial University of Newfoundland, St. John's, A1B 3X5, Canada}
\address{\hypertarget{corr}{*}Corresponding author. \email{tzou@mun.ca}}


\abstract{Accurate inertial odometry is critical for autonomous systems but suffers from unbounded drift caused by the integration of sensor noise and biases. In approaches such as TLIO, neural pseudo-measurements are trained independently of their downstream filter. This paper introduces Differentiable Filtering for Inertial Odometry (DFIO), a learning framework that embeds the update stages of an error-state Kalman filter directly into a computational graph. Following standalone pre-training, DFIO jointly
optimizes the displacement and covariance output heads using
posterior-position supervision through differentiable measurement
updates. The proposed system is validated on public pedestrian motion datasets featuring both calibrated and uncalibrated inertial measurements. Experimental results on a pedestrian dataset demonstrate that DFIO reduces mean absolute and one-second relative translation errors by 13.2\% and 39.6\% relative to the TLIO baseline, and by 7.7\% and 17.4\% relative to its pre-trained counterpart. Furthermore, evaluations on uncalibrated smartphone inertial data indicate that the proposed method effectively mitigates translational drift and maintains trajectory alignment under highly noisy sensor conditions. Source code will be made publicly available.}

\maketitle

\section{Introduction}
	
Reliable estimates of position and orientation are essential for a multitude of engineering applications, including robot navigation~\cite{barfoot2024state}, autonomous-vehicle localization~\cite{kuutti2018survey}, the registration of virtual content in augmented reality~\cite{marchand2015pose}, and pedestrian motion tracking~\cite{chen2021data, zhang2023idt}, to name a few. For mobile platforms, inertial measurement units (IMUs) offer a compact, low-cost source of high-rate motion information. Because they measure angular velocity and specific force without observing the surrounding scene, they remain useful when external positioning signals or visual features are unavailable~\cite{kok2017inertial,liu2020tlio}. These advantages make inertial odometry (IO) an important component of autonomous navigation, but obtaining accurate trajectories from inertial measurements alone remains challenging.

\begin{figure}[tbp]
    \centering
    \includegraphics[width=0.56\linewidth]{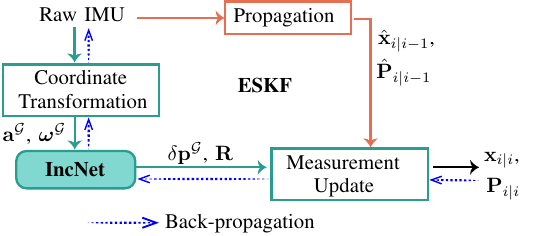}
    \caption{Overview of DFIO. Raw IMU measurements drive analytical ESKF propagation and are transformed into a local gravity-aligned frame $\mathcal G$ for neural measurement model IncNet. The specific force $\mathbf a^{\mathcal G}$ and angular velocity $\boldsymbol\omega^{\mathcal G}$ are processed by IncNet to predict a displacement increment $\delta\mathbf p^{\mathcal G}$ and its covariance $\mathbf R$ for the ESKF measurement update. The pairs $(\hat{\mathbf x}_{i|i-1},\hat{\mathbf P}_{i|i-1})$ and $(\mathbf x_{i|i},\mathbf P_{i|i})$ denote the prior and posterior state estimates and error-state covariances, respectively, for measurement windows $i$. After standalone pretraining, posterior-position supervision through the differentiable update refines the displacement and covariance heads of IncNet. Blue dotted arrows indicate backpropagation during training}
    \label{fig:flowchart1}
\end{figure}

Strapdown inertial navigation first integrates angular velocity to estimate orientation, then uses that orientation to transform specific force into a navigation frame and recover velocity and position~\cite{savage1998strapdown}. Biases, measurement noise, and orientation errors therefore affect not only individual estimates but also the subsequent trajectory, causing drift to accumulate over time~\cite{kok2017inertial}. Calibration can reduce systematic errors, including nonlinear scale-factor effects~\cite{zhang2021low}, but residual errors and stochastic noise still propagate through the integration~\cite{burnett2025imu}.

A common way to control this drift is to introduce complementary observations. Chen et al.~\cite{chen2021data} combined learned pedestrian motion increments with Bluetooth localization in a particle filter. In robotic navigation, Alkhawaja et al.~\cite{alkhawaja2023low} fused depth-based localization with inertial estimates using cascade-correlation networks, while Zhang et al.~\cite{zhang2025tightly} combined learned optical-flow tracking with visual--inertial optimization under changing illumination. These approaches demonstrate how additional observations can complement inertial integration. When such observations are unavailable, however, an IMU-only estimator must instead draw on information about the motion itself.

Learning-based IO provides a way to extract this information from data~\cite{chen2024deep}. Methods such as IONet~\cite{chen2018ionet} and RoNIN~\cite{herath2020ronin} learn motion increments or velocities from inertial sequences and compose them into trajectories. Their predictions capture regularities that are difficult to express through sensor calibration alone, but prediction errors can still accumulate during trajectory reconstruction. Hybrid methods therefore combine learned motion measurements with analytical IMU propagation: TLIO~\cite{liu2020tlio} supplies displacement measurements and their covariances to an extended Kalman filter (EKF), while DIVE~\cite{bajwa2024dive} uses learned velocity measurements in an on-manifold EKF. The filter balances these learned motion priors against its propagated estimate according to their uncertainties.

When learned predictions are fused with an inertial prior, training can also account for their effect on the resulting state estimate. In TLIO and DIVE, the measurement networks are trained against reference displacement or velocity targets. Such supervision improves the local predictions, but does not directly optimize the state estimate obtained after the filter combines them with its prior. Differentiable filtering provides a means of bringing that downstream objective into training~\cite{haarnoja2016backprop,kloss2021train}. This idea is already relevant to inertial navigation: AI-IMU Dead-Reckoning~\cite{brossard2020ai} and the pedestrian method of Wang et al.~\cite{wang2020pedestrian} train covariance adapters using errors in the filter's estimated motion. These methods learn how strongly to trust prescribed motion constraints, but they need to construct their pseudo-measurements from specific assumptions about the motion. This motivates investigating whether filter-output supervision can refine pseudo-measurements learned directly from IMU data.

We address this question with Differentiable Filtering for Inertial Odometry (DFIO). As shown in Fig.~\ref{fig:flowchart1}, DFIO combines an error-state Kalman filter (ESKF) with IncNet, a temporal network that predicts three-dimensional displacement increments and diagonal measurement covariances in a local gravity-aligned frame. The network is first trained using displacement supervision. Its displacement and covariance heads are then refined together using posterior-position supervision through differentiable filter updates, while the pretrained temporal encoder is held fixed. The propagation model retains the deterministic IMU kinematics; the learnable quantities are the neural measurement predictions and their uncertainties. This allows the training objective to assess those predictions through their effect on the estimated trajectory.

The contributions of this work are:
\begin{itemize}
    \item A learned inertial measurement model that combines temporal residual features and self-attention to predict three-dimensional displacement increments and measurement covariances for an ESKF.
    \item A two-stage training strategy that jointly refines the displacement and covariance heads using filter posterior-position feedback, while retaining the analytical IMU propagation model of the ESKF.
    \item An evaluation on held-out TLIO pedestrian trajectories, with comparisons and ablations examining the effects of displacement and covariance refinement. An additional evaluation on a subset of the OxIOD dataset~\cite{chen2020deep} examines performance with noisy, uncalibrated smartphone IMU measurements.
\end{itemize}

\section{Related Work}
\label{sec:related_work}

\subsection{Learning-Assisted Inertial Odometry}

Learning has been applied at different levels of inertial estimation, from correcting state estimates to predicting motion directly~\cite{chen2024deep, al2019deep}. For translational odometry, a related motivation is to learn the relationship between an inertial sequence and the motion that generated it, thereby reducing reliance on unconstrained integration.

IONet~\cite{chen2018ionet} uses a recurrent architecture based on long short-term memory (LSTM)~\cite{hochreiter1997long} to predict windowed displacement magnitudes and heading changes, which are composed into pedestrian trajectories. RoNIN~\cite{herath2020ronin} develops this direction by comparing residual networks~\cite{he2016deep}, LSTMs, and temporal convolutional networks~\cite{bai2018empirical} for horizontal motion prediction and trajectory reconstruction. To capture richer context within inertial sequences, CTIN~\cite{rao2022ctin} combines a residual encoder with self-attention for motion prediction. To optimize computational cost, the OxIOD study~\cite{chen2020deep} examines lightweight learned inertial navigation and on-device inference. IMUNet~\cite{zeinali2024imunet} combines a MobileResNet-based noise-estimation stage with a learned velocity-estimation stage, jointly supervised using reference velocities. Together, these studies show how learned motion models and sensor corrections can supplement conventional inertial integration.

Fusion provides a further means of using the learned motion information. TLIO~\cite{liu2020tlio} predicts three-dimensional relative displacements and their covariances and incorporates them in a stochastic-cloning EKF. The augmented state allows relative measurements between past and current poses to update the estimated motion. Building on the TLIO filtering framework, EqNIO~\cite{jayanth2025eqnio} introduces gravity-preserving equivariant representations for displacement and covariance prediction. DIVE~\cite{bajwa2024dive} predicts velocity and its covariance from IMU-derived histories and uses an on-manifold EKF for quadrotor state estimation. In TLIO and DIVE, the neural measurement models are trained against local reference targets before their predictions are used in the filter. This separation motivates examining whether training through the filter can further improve the final estimates.

A complementary approach starts from motion constraints and learns how much confidence to assign to them. AI-IMU Dead-Reckoning~\cite{brossard2020ai} uses approximately zero lateral and vertical velocities in the vehicle frame as pseudo-measurements in an invariant EKF, with a neural adapter predicting time-varying covariances. Its adapter and selected filter parameters are trained using filter trajectory errors. Wang et al.~\cite{wang2020pedestrian} extend covariance adaptation to pedestrian velocity pseudo-observations and train the adapter by comparing filter-estimated body-frame velocities with reference velocities. Thus, filter-output supervision is already established in inertial navigation. However, these pseudo-measurements rely on specific motion assumptions, and may not generalize to arbitrary movements. In contrast, DFIO learns pseudo-measurements directly from IMU sequences and refines them jointly with their covariances through the filter update.

\subsection{Differentiable Filtering for State Estimation}

Training a learned measurement model through its estimator connects prediction quality to the state-estimation objective. Backprop KF~\cite{haarnoja2016backprop} demonstrated this principle by placing a neural observation model and Kalman-filter recurrence in a computation graph, with experiments on visual tracking and the KITTI visual-odometry benchmark~\cite{geiger2012we}. Related formulations retain the structure of particle filters~\cite{jonschkowski2018differentiable,karkus2018particle} and histogram filters~\cite{jonschkowski2016endtoend} while learning models for their inference objectives. The common idea is to preserve the estimator's algorithmic structure while adapting its learnable components to the final task.

Subsequent work has explored which components benefit from learning. DynaNet~\cite{chen2021dynanet} learns time-varying latent dynamics within a neural Kalman model, while KalmanNet~\cite{revach2022kalmannet} uses a recurrent network to estimate the Kalman gain. Liu et al.~\cite{liu2023enhancing} develop a differentiable ensemble Kalman filter with learned stochastic dynamics and evaluate it on visual odometry. Alongside these architectural choices, uncertainty modeling and training stability are central considerations. Kloss et al.~\cite{kloss2021train} compare several differentiable filters and training strategies, finding particular benefits from learning noise models through the filter, while also showing that the value of joint training depends on the component and task. These findings make component-wise evaluation important for identifying which parts of a learned estimator benefit from supervision through the filter.

Differentiable filtering also provides a way to combine known dynamics with learned sensing. Li and Waslander~\cite{li2020towards} combine analytical IMU propagation with learned visual relative-pose measurements and covariances, using both intermediate measurement supervision and a final state-estimation objective. For heterogeneous sensing, Lee et al.~\cite{lee2020multimodal} study differentiable fusion of visual, haptic, and proprioceptive information, including learned crossmodal weighting. These studies illustrate how estimator structure can organize learning around the final state estimate while accommodating different choices of physical and learned models.

DFIO builds on this line of work for measurements derived solely from IMU data. Its focus is the joint refinement of learned displacement means and covariances using posterior-position feedback from ESKF updates. The experiments examine the benefit of this refinement and the individual effects of displacement and covariance adaptation in pedestrian IO.	
	
\section{Preliminaries}

\subsection{Notation and Coordinate Frames}

Throughout this paper, scalars are written in italics (e.g., $a$), vectors in bold lowercase (e.g., $\mathbf a$), and matrices in bold uppercase (e.g., $\mathbf A$). Coordinate frames are represented by calligraphic capitals (e.g., $\mathcal{A}$). The primary frames are:
\begin{itemize}
    \item The world frame $\mathcal W$, an inertial reference whose $z$-axis points opposite to gravity.
    \item The body frame $\mathcal B$, rigidly attached to the IMU sensor.
    \item The local gravity-aligned frame $\mathcal G_i$, associated with measurement window $i$. Its vertical axis is parallel to that of $\mathcal W$, and its horizontal orientation is determined by the anchor yaw $\hat\psi_{i,0}$, as defined in Section~\ref{sec:input_parameterization}.
\end{itemize}

A superscript identifies the frame in which a vector is expressed. World-frame superscripts are omitted for brevity. The body-to-world rotation $\mathbf{C}_k\in SO(3)$ maps a body-frame vector to the world frame. 
A tilde denotes a raw IMU measurement, a hat denotes an estimate, and superscripts $-$ and $+$ distinguish the prior and posterior estimates at a measurement update.

The index $k$ denotes an IMU sampling step, $i$ denotes a measurement window, and $\tau$ denotes a sampling time within a window. The IMU sampling interval is $\delta t_k=t_{k+1}-t_k$. 

	
	\subsection{IMU Measurement Model}\label{sec:imu_model}
	
	An IMU measures the angular velocity and specific force in its body frame $\mathcal{B}$. At a discrete time step $t_k$, the raw measurements from the gyroscope, $\tilde{\boldsymbol\omega}_{k}^{\mathcal B}$, and accelerometer, $\tilde{\mathbf a}_{k}^{\mathcal B}$, are modeled as:
	\begin{subequations}\label{eq:imu_model}
		\begin{align}
		\tilde{\boldsymbol\omega}_{k}^{\mathcal B} &= \boldsymbol\omega_{k}^{\mathcal B} + \mathbf{b}_{g,k}^{\mathcal B} + \boldsymbol\eta_{g,k}^{\mathcal B}, \\
		\tilde{\mathbf a}_{k}^{\mathcal B} &= \mathbf{a}_{k}^{\mathcal B} - \mathbf{C}_{k}^{\top}\mathbf{g} + \mathbf{b}_{a,k}^{\mathcal B} + \boldsymbol\eta_{a,k}^{\mathcal B},
		\end{align}
	\end{subequations}
	where $\boldsymbol\omega_{k}^{\mathcal B}$ is the true angular velocity and $\mathbf{a}_{k}^{\mathcal B}$ is the true kinematic acceleration, both expressed in the body frame. Given the world-frame gravity vector $\mathbf{g}$, the accelerometer measures the specific force $\mathbf{a}_{k}^{\mathcal B} - \mathbf{C}_{k}^{T}\mathbf{g}$. The measurements are corrupted by slowly varying biases, $\mathbf{b}_{g,k}^{\mathcal B}$ and $\mathbf{b}_{a,k}^{\mathcal B}$, and are subject to zero-mean, white Gaussian noise, $\boldsymbol\eta_{g,k}^{\mathcal B} \sim \mathcal{N}(\mathbf{0}, \sigma_g^2 \mathbf{I})$ and $\boldsymbol\eta_{a,k}^{\mathcal B} \sim \mathcal{N}(\mathbf{0}, \sigma_a^2 \mathbf{I})$. The biases are modeled as random walks:
	\begin{subequations}\label{eq:bias_rw}
		\begin{align}
		\mathbf{b}_{g,k+1}^{\mathcal B} &= \mathbf{b}_{g,k}^{\mathcal B} + \boldsymbol\nu_{g,k}^{\mathcal B}, \\
		\mathbf{b}_{a,k+1}^{\mathcal B} &= \mathbf{b}_{a,k}^{\mathcal B} + \boldsymbol\nu_{a,k}^{\mathcal B},
		\end{align}
	\end{subequations}
	where the driving noise terms are $\boldsymbol\nu_{g,k}^{\mathcal B} \sim \mathcal{N}(\mathbf{0}, \sigma_{bg}^2 \mathbf{I})$ and $\boldsymbol\nu_{a,k}^{\mathcal B} \sim \mathcal{N}(\mathbf{0}, \sigma_{ba}^2 \mathbf{I})$.

\subsection{Error-State Kalman Filter Framework}
\label{sec_ekf_model}

The ESKF propagates a nominal state using nonlinear IMU kinematics and estimates small perturbations around that state using linearized error dynamics. The estimated perturbations are then injected into the nominal state to correct the trajectory.

\subsubsection{State Definition}
The true state at time $t_k$ is
\begin{equation}
\mathbf{x}_k=
\left(\mathbf p_k,\mathbf v_k,\mathbf C_k,
\mathbf b_{a,k}^{\mathcal B},\mathbf b_{g,k}^{\mathcal B}\right),
\label{eq:true_state}
\end{equation}
where $\mathbf p_k,\mathbf v_k\in\mathbb R^3$ are world-frame position and velocity, $\mathbf C_k\in SO(3)$ is the body-to-world rotation, and $\mathbf{b}_{a,k}^{\mathcal B}, \mathbf{b}_{g,k}^{\mathcal B} \in \mathbb{R}^3$ are the accelerometer and gyroscope biases. Its nominal estimate is denoted by $\hat{\mathbf x}_k$. The associated 15-dimensional error state is
\begin{equation}
\delta\mathbf x_k=
\begin{bmatrix}
\delta\mathbf p_k^\top &
\delta\mathbf v_k^\top &
\boldsymbol\phi_k^\top &
(\delta\mathbf b_{a,k}^{\mathcal B})^\top &
(\delta\mathbf b_{g,k}^{\mathcal B})^\top
\end{bmatrix}^{\top}.
\label{eq:error_state}
\end{equation}
Position, velocity, and bias errors are additive, for example $\mathbf p_k=\hat{\mathbf p}_k+\delta\mathbf p_k$. Attitude uses a left-multiplicative perturbation,
\begin{equation}
\mathbf C_k=\exp_{SO(3)}(\boldsymbol\phi_k)\hat{\mathbf C}_k,
\label{eq:attitude_error}
\end{equation}
We denote this combination of additive and rotational perturbations by $\mathbf x_k=\hat{\mathbf x}_k\boxplus\delta\mathbf x_k$.
\subsubsection{State Propagation}
\label{sec:ekf_propag}
Between measurement updates, the nominal state is propagated at the IMU sampling frequency. Over one interval $\delta t_k$, the discrete-time kinematics are
\begin{subequations}\label{eq:nominal_prop}
\begin{align}
\hat{\mathbf C}_{k+1}^{-}
    &=\hat{\mathbf C}_{k}^{+}
      \exp_{SO(3)}\!\left[
      (\tilde{\boldsymbol\omega}_{k}^{\mathcal B}
      -\hat{\mathbf b}_{g,k}^{\mathcal B,+})\delta t_k\right],\\
\hat{\mathbf v}_{k+1}^{-}
    &=\hat{\mathbf v}_{k}^{+}+\hat{\mathbf a}_{k}^{\mathcal W}\delta t_k,\\
\hat{\mathbf p}_{k+1}^{-}
    &=\hat{\mathbf p}_{k}^{+}+\hat{\mathbf v}_{k}^{+}\delta t_k
      +\tfrac12\hat{\mathbf a}_{k}^{\mathcal W}\delta t_k^2,
\end{align}
\end{subequations}
where $\hat{\mathbf a}_{k}^{\mathcal W}=
\hat{\mathbf C}_{k}^{+}(\tilde{\mathbf a}_{k}^{\mathcal B}
-\hat{\mathbf b}_{a,k}^{\mathcal B,+})+\mathbf g$ is the estimated world-frame kinematic acceleration. The nominal bias estimates remain unchanged during prediction because the random-walk increments have zero mean:
\begin{equation}
\hat{\mathbf b}_{a,k+1}^{\mathcal B,-}
=\hat{\mathbf b}_{a,k}^{\mathcal B,+},\qquad
\hat{\mathbf b}_{g,k+1}^{\mathcal B,-}
=\hat{\mathbf b}_{g,k}^{\mathcal B,+}.
\label{eq:nominal_bias_prop}
\end{equation}
Their uncertainty is included in the covariance propagation,
\begin{equation}
\mathbf P_{k+1}^{-}
=\mathbf F_k\mathbf P_k^{+}\mathbf F_k^\top+\mathbf Q_k,
\label{eq:cov_time}
\end{equation}
where $\mathbf{P}_k$ describes the error-state uncertainty, $\mathbf F_k$ is the discrete error-state transition matrix and $\mathbf{Q}_{k}$ is the discretized process noise covariance matrix. 

	
\subsubsection{Measurement Correction and Error Reset}
For a measurement or pseudo-measurement $\mathbf z_k$ with model $\mathbf z_k=h_k(\mathbf x_k)+\boldsymbol\epsilon_k$, assume $\boldsymbol\epsilon_k\sim\mathcal N(\mathbf0,\mathbf R_k)$. The standard update further assumes that this error is uncorrelated with the prior state error. Linearization about $\hat{\mathbf x}_k^{-}$ gives
\begin{equation}
\mathbf H_k=
\left.\frac{\partial h_k(\hat{\mathbf x}_k^{-}\boxplus\delta\mathbf x)}
{\partial\delta\mathbf x}\right|_{\delta\mathbf x=\mathbf0}.
\label{eq:generic_measurement_jacobian}
\end{equation}
The innovation, its covariance, and the Kalman gain are
\begin{subequations}\label{eq:ekf_update}
\begin{align}
\mathbf y_k&=\mathbf z_k-h_k(\hat{\mathbf x}_k^{-}),
\label{eq:innovation}\\
\mathbf S_k&=\mathbf H_k\mathbf P_k^{-}\mathbf H_k^\top+\mathbf R_k,\\
\mathbf K_k&=\mathbf P_k^{-}\mathbf H_k^\top\mathbf S_k^{-1}.
\end{align}
\end{subequations}
The estimated error correction and covariance are computed as
\begin{subequations}\label{eq:ekf_state_update_prelim}
\begin{align}
\widehat{\delta\mathbf x}_k&=\mathbf K_k\mathbf y_k,\\
\bar{\mathbf P}_k&=(\mathbf I -\mathbf K_k\mathbf H_k)\mathbf P_k^{-}.
\end{align}
\end{subequations}
The correction is then injected through $\hat{\mathbf x}_k^{+}
=\hat{\mathbf x}_k^{-}\boxplus\widehat{\delta\mathbf x}_k$.
After nominal-state injection, the general covariance reset is
\begin{equation}
\mathbf P_k^+
=
\mathbf G_k\bar{\mathbf P}_k\mathbf G_k^\top,
\end{equation}
where $\mathbf G_k$ is the Jacobian of the error-state reset.

\section{Methodology}
\label{sec:methodology}

As illustrated in Fig.~\ref{fig:flowchart1}, the proposed DFIO framework is formulated as a differentiable state estimation graph. Rather than treating neural inertial odometry as an independent regression task, DFIO tightly couples a deep temporal measurement encoder (\textit{IncNet}) with an ESKF, enabling the joint refinement of displacement predictions and their covariances through the filter updates.


\subsection{Learned Measurement Model: IncNet}
\label{sec:IncNet}

For measurement window $i$, containing $n$ IMU samples at times $t_{i,0},\ldots,t_{i,n-1}$, IncNet regresses a 3D relative position increment $\widehat{\Delta\mathbf p}_i^{\mathcal G_i}\in\mathbb R^3$ alongside an associated heteroscedastic measurement covariance $\mathbf R_i\in\mathbb R^{3\times3}$.

\subsubsection{Input Parameterization}
\label{sec:input_parameterization}
Following the design principles of TLIO~\cite{liu2020tlio} and DIVE~\cite{bajwa2024dive}, IncNet operates entirely within a local, gravity-aligned coordinate frame $\mathcal G_i$ established at the window anchor $t_{i,0}$. This removes dependence on arbitrary global headings from the network representation.

The raw body-frame IMU samples ($\tilde{\mathbf a}_\tau^{\mathcal B},\tilde{\boldsymbol\omega}_\tau^{\mathcal B}$ for $\tau\in[t_{i,0},t_{i,n-1}]$) are first transformed into the world frame $\mathcal{W}$:
\begin{equation}\label{eq:imu_world_transform}
	\tilde{\mathbf a}_\tau =  \hat{\mathbf C}(\tau)\,\tilde{\mathbf{a}}_\tau^{\mathcal{B}}, \quad
	\tilde{\boldsymbol\omega}_\tau =  \hat{\mathbf C}(\tau)\,\tilde{\boldsymbol{\omega}}_\tau^{\mathcal{B}}.
\end{equation}
In both training stages and at inference, the rotations $\hat{\mathbf C}(\tau)$ are obtained by gyroscope integration initialized with the reference orientation at the start of each sequence. The resulting world-frame IMU channels are normalized using per-axis robust scalers fitted on the
training set. The network input for window $i$ is
\begin{equation}
\mathbf u_i^{\mathcal G_i}
=
\operatorname{blkdiag}(\mathbf A_i,\mathbf A_i)
\mathcal S(\mathbf u_i^{\mathcal W}),
\qquad
\mathbf A_i=\mathbf C_z^\top(\hat\psi_{i,0}),
\label{eq:imu_gla_transform}
\end{equation}
where $\mathbf u_i^{\mathcal W}$ stacks the world-frame
specific-force and angular-velocity channels, and $\mathcal S$
denotes their normalization transforms. During standalone pre-training (Phase~1 in Section~\ref{sec:training_loss}), synthetic random yaw rotations $\hat\psi_{i,0}\sim\mathcal U(-\pi,\pi)$ are applied consistently to both the IMU inputs and their displacement targets. During fine-tuning stage (Phase~2  in Section~\ref{sec:training_loss}) and inference,
the ESKF provides the estimated anchor yaw $\hat\psi_{i,0}$ for
this transformation.


\subsubsection{Network Architecture and Outputs}

The network extends the 1D temporal residual architecture proposed in~\cite{liu2020tlio} by incorporating self-attention mechanisms. The input sequence passes through an initial $7\times1$ convolutional stem (stride 2) for feature downsampling, followed by four progressive residual stages ($64\rightarrow128\rightarrow256\rightarrow512$ channels). Each stage contains two depthwise-separable residual blocks coupled with a Multi-Head Self-Attention (MHSA) module operating across the temporal axis.

A global average pooling layer aggregates the final temporal feature map into a fixed 512-dimensional context vector. This vector is fed into two parallel, fully-connected output heads:
a displacement head predicting
$\widehat{\Delta\mathbf p}_i^{\mathcal G_i}$, and an uncertainty
head regressing a 3-coefficient parameter vector
$\boldsymbol\ell_i\in\mathbb R^3$.
Assuming the measurement noise is uncorrelated across the spatial
axes, the covariance is diagonal, with
$\boldsymbol\ell_i$
parameterizing the log standard deviations:
\begin{equation}
\mathbf R_i(\boldsymbol\ell_i)
=
\operatorname{diag}\left(
e^{2\ell_{i,x}},
e^{2\ell_{i,y}},
e^{2\ell_{i,z}}
\right).
\label{eq:cov_param}
\end{equation}
\subsection{Displacement Measurement Update}

For window $i$, the saved anchor position and yaw are held fixed
in the linearized measurement update. The displacement observation
and innovation are
\begin{align}
h_i(\hat{\mathbf x}_i^-)
&=
\mathbf A_i
(\hat{\mathbf p}_i^--\hat{\mathbf p}_{i,0}),
\label{eq:dfio_displacement_observation}\\
\mathbf y_i^{\mathcal G_i}
&=
\widehat{\Delta\mathbf p}_i^{\mathcal G_i}
-
h_i(\hat{\mathbf x}_i^-).
\label{eq:dfio_displacement_innovation}
\end{align}
For the error-state ordering in~\eqref{eq:error_state}, the
Jacobian is
\begin{equation}
\mathbf H_i=
\begin{bmatrix}
\mathbf A_i & \mathbf 0_{3\times12}
\end{bmatrix}.
\label{eq:dfio_displacement_jacobian}
\end{equation}
The update uses~\eqref{eq:ekf_update} with the predicted covariance
$\mathbf R_i$.

\subsection{Two-Stage Curriculum Training}
\label{sec:training_loss}

Optimizing a tightly coupled neural-filtering system from random initialization can be unstable when the initial displacement and covariance predictions are inaccurate. Optimization is therefore structured into a two-stage curriculum.

\subsubsection{Phase 1: Standalone Measurement Pre-training}

The initial phase isolates the \textit{IncNet} module, bypassing the filter to establish a competent baseline displacement estimator. Following the MSE-to-NLL training schedule used in~\cite{liu2020tlio,bajwa2024dive}, the network is first trained for 10 epochs minimizing the Mean Squared Error (MSE) against the ground-truth window displacement $\Delta\mathbf p_i^{\mathcal G_i,\mathrm{gt}}$:
\begin{equation}
\mathcal L_{\mathrm{MSE}}
=\frac1B\sum_{i=1}^{B}
\left\|
\widehat{\Delta\mathbf p}_i^{\mathcal G_i}
-\Delta\mathbf p_i^{\mathcal G_i,\mathrm{gt}}
\right\|_2^2,
\label{eq:phase1_mse}
\end{equation}
where $B$ is the batch size and each target is expressed in the same coordinates as its prediction. Once the displacement estimates stabilize, the loss transitions to the marginal NLL for the remaining epochs:
\begin{equation}
\mathcal L_{\mathrm{NLL}}
=\frac{1}{2B}\sum_{i=1}^{B}
\left(
\mathbf e_i^\top\mathbf R_i^{-1}\mathbf e_i
+\ln\det\mathbf R_i
\right),
\label{eq:phase1_nll}
\end{equation}
where $\mathbf e_i
=\Delta\mathbf p_i^{\mathcal G_i,\mathrm{gt}}
-\widehat{\Delta\mathbf p}_i^{\mathcal G_i}$.
At the conclusion of Phase~1, \textit{IncNet} predicts local motion and its input-dependent uncertainty, but has not yet been optimized using filter-output supervision.

\subsubsection{Phase 2: Filter-Coupled Refinement}
In Phase~2, \textit{IncNet} is embedded into the closed-loop ESKF computation graph. To preserve the features learned in Phase~1, the temporal encoder is frozen and held in evaluation mode, while estimation errors backpropagated through the filter update only the linear output projection heads. 

Gradients from posterior-position supervision pass through the error-state correction and covariance update~\eqref{eq:ekf_state_update_prelim}. Through the measurement innovation $\mathbf y_i^{\mathcal G_i}$ and Kalman gain $\mathbf K_i$ in~\eqref{eq:ekf_update}, these gradients reach the predicted displacement $\widehat{\Delta\mathbf p}_i^{\mathcal G_i}$ and covariance $\mathbf R_i$. This allows the two output heads to be refined according to their effect on the posterior position and covariance. 

At each measurement update $i$, the coupled system minimizes
\begin{equation}
\mathcal L_{\mathrm{Phase2}}
=\mathcal L_{\mathrm{pos\text{-}NLL}}
+\lambda_{\mathrm{NEES}}\mathcal L_{\mathrm{NEES}},
\label{eq:total_phase2}
\end{equation}
where $\lambda_{\mathrm{NEES}}=0.1$. The primary objective supervises the posterior positions $\hat{\mathbf p}_i^+$ and their $3\times3$ marginal covariances $\mathbf P_{p,i}^+$ against reference positions:
\begin{equation}
\mathcal L_{\mathrm{pos\text{-}NLL}}
=\frac{1}{2}
\left[
\mathbf e_{p,i}^\top
(\mathbf P_{p,i}^{+})^{-1}\mathbf e_{p,i}
+\ln\det\mathbf P_{p,i}^{+}
\right],
\end{equation}
where $\mathbf e_{p,i}
=\mathbf p_i^{\mathrm{gt}}-\hat{\mathbf p}_i^{+}$ and
$\mathbf P_{p,i}^{+}$ is the position block of $\mathbf P_i^{+}$.

To additionally regularize the scale of covariance-normalized posterior-position errors, we use a penalty based on the position normalized estimation error squared (NEES),
\begin{equation}
q_{p,i}=\mathbf e_{p,i}^\top
(\mathbf P_{p,i}^+)^{-1}\mathbf e_{p,i}.
\end{equation}
The expected NEES under a correctly specified zero-mean Gaussian position-error model is the position dimension $d_p=3$. Using this value as a reference, we define
\begin{equation}
\mathcal L_{\mathrm{NEES}}
=\left[
\operatorname{clip}\!\left(
\ln\!\left(\frac{q_{p,i}}{d_p}\right),-7,7
\right)\right]^2.
\label{eq:phase2_nees}
\end{equation}
The squared logarithmic ratio penalizes multiplicative deviations above and below the reference value, while clipping limits the contribution of extreme values. This auxiliary term is a heuristic covariance-scale regularizer rather than a statistical consistency test or a guarantee of calibrated uncertainty. The combined loss is averaged over the trajectories processed in parallel. Algorithm~\ref{alg_main_pathway} summarizes DFIO filtering, and Algorithm~\ref{alg_phase2_training} summarizes Phase~2 refinement using the same filtering operations.
\subsection{Implementation Details}
	
DFIO was implemented in PyTorch and trained on an NVIDIA GeForce RTX 3070 GPU. To prepare inputs for the IncNet module, raw IMU data is processed into windows of $n=100$ samples (approximately 0.5~s at 200~Hz). In Phase~1, the network is optimized using Adam with an initial learning rate of $10^{-4}$ and a \texttt{ReduceLROnPlateau} scheduler (patience 5, decay factor 0.5). In Phase~2, the unfrozen IncNet linear projection heads are updated at a learning rate of $10^{-3}$. During Phase 2 training, the position, velocity, and orientation at the start of each loader segment are set to their reference values; bias estimates and covariance are carried between consecutive segments of the same trajectory. During inference, the estimated state is carried forward without these reference resets.
\begin{algorithm}[t]
\caption{DFIO Filtering Process}
\label{alg_main_pathway}
\begin{algorithmic}[1]
\Require Body-frame IMU sequence $\mathbf u^{\mathcal B}$,
initial state $\mathbf x_0$, covariance $\mathbf P_0$, IMU noise parameters, sampling interval $\delta t$,
trained IncNet, and saved input/output scalers
$\mathcal S,\mathcal T$
\Ensure State history $\mathscr X$, covariance history $\mathscr P$

\State $\hat{\mathbf x}_0^+\gets\mathbf x_0$,
       $\mathbf P_0^+\gets\mathbf P_0$
\State $\mathscr X\gets\varnothing$,
       $\mathscr P\gets\varnothing$

\State $\mathbf u^{\mathcal W}\gets
       \operatorname{GyroPreprocess}
       (\mathbf u^{\mathcal B},\hat{\mathbf C}_0^+)$
       using~\eqref{eq:imu_world_transform}
       \label{algline:preprocess}
\State $\bar{\mathbf u}^{\mathcal W}
       \gets\mathcal S(\mathbf u^{\mathcal W})$
\State Form $m$ windows of $n$ samples with stride $n-1$

\For{$i=1,\ldots,m$}
    \State $(\hat{\mathbf x}_{i,0},\mathbf P_{i,0})
           \gets
           (\hat{\mathbf x}_{i-1}^+,\mathbf P_{i-1}^+)$

    \State Save anchor
           $(\hat{\mathbf p}_{i,0},\hat{\mathbf C}_{i,0})$
           from $\hat{\mathbf x}_{i,0}$
           \label{algline:filter_window_start}

    \State $\hat\psi_{i,0}\gets
           \operatorname{yaw}(\hat{\mathbf C}_{i,0})$,
           $\mathbf A_i\gets
           \mathbf C_z^\top(\hat\psi_{i,0})$

    \State $\mathbf u_i^{\mathcal G_i}\gets
           \operatorname{blkdiag}(\mathbf A_i,\mathbf A_i)
           \bar{\mathbf u}_i^{\mathcal W}$

    \State $(\boldsymbol\mu_i,\boldsymbol\ell_i)
           \gets
           \operatorname{IncNet}(\mathbf u_i^{\mathcal G_i})$

    \For{$k=0,\ldots,n-2$}
        \State $(\hat{\mathbf x}_{i,k+1},\mathbf P_{i,k+1})
               \gets
               f(\hat{\mathbf x}_{i,k},\mathbf P_{i,k},
               \mathbf u_{i,k}^{\mathcal B},\delta t)$
    \EndFor

    \State $(\hat{\mathbf x}_i^-,\mathbf P_i^-)
           \gets
           (\hat{\mathbf x}_{i,n-1},\mathbf P_{i,n-1})$

    \State
           $(\widehat{\Delta\mathbf p}_i^{\mathcal G_i},
           \mathbf R_i)
           \gets
           \operatorname{MetricOutputs}
           (\boldsymbol\mu_i,\boldsymbol\ell_i;\mathcal T)$

    \State Floor the diagonal entries of $\mathbf R_i$
           at $10^{-8}\,\mathrm{m}^2$

    \State Compute $\mathbf y_i^{\mathcal G_i}$
           and $\mathbf H_i$
           using~\eqref{eq:dfio_displacement_innovation}
           and~\eqref{eq:dfio_displacement_jacobian}

    \State $(\hat{\mathbf x}_i^+,\mathbf P_i^+)
           \gets
           \operatorname{ESKFCorrect}
           (\hat{\mathbf x}_i^-,\mathbf P_i^-,
           \mathbf y_i^{\mathcal G_i},\mathbf H_i,\mathbf R_i)$
           \label{algline:filter_window_end}

    \State Append $\hat{\mathbf x}_i^+$ to $\mathscr X$
           and $\mathbf P_i^+$ to $\mathscr P$
\EndFor

\State \Return $\mathscr X,\mathscr P$
\end{algorithmic}
\end{algorithm}
\begin{algorithm}[t]
\caption{Filter-Coupled Refinement}
\label{alg_phase2_training}
\begin{algorithmic}[1]
\Require Training and validation trajectories with reference states, Phase~1 IncNet, and filter initialization
\Ensure Refined IncNet
\State Freeze the temporal encoder and hold it in evaluation mode
\State Preprocess each trajectory as in Algorithm~\ref{alg_main_pathway}, line~\ref{algline:preprocess}
\State Initialize Adam for the two output heads, with parameters $\boldsymbol\theta_{\mathrm h}$
\For{each training epoch}
    \State Initialize the state and covariance independently for each trajectory
    \For{each single-window batch $i$ in temporal order}
        \State $(\hat{\mathbf x}_{i,0},\mathbf P_{i,0})\gets$ carried state and covariance
        \State Replace the initial position, velocity, and orientation with their reference values
        \State Obtain $(\hat{\mathbf x}_i^+,\mathbf P_i^+)$ using Algorithm~\ref{alg_main_pathway}, lines~\ref{algline:filter_window_start}--\ref{algline:filter_window_end}
        \State Compute the batch-mean loss $\overline{\mathcal L}_i$ using~\eqref{eq:total_phase2}
        \State Clear gradients and backpropagate $\overline{\mathcal L}_i$ to $\boldsymbol\theta_{\mathrm h}$
        \State Clip the gradient norm to $5$ and update $\boldsymbol\theta_{\mathrm h}$ using Adam
        \State Carry detached copies of $\hat{\mathbf x}_i^+$ and $\mathbf P_i^+$ to the next window
    \EndFor
    \State Evaluate validation loss; retain the best IncNet and apply learning-rate and stopping criteria
\EndFor
\State \Return IncNet with the lowest validation loss
\end{algorithmic}
\end{algorithm}
\section{Experimental Results}
\label{sec:experimental_results}

\subsection{Experimental Setup}

\subsubsection{Datasets}
We evaluate DFIO on two public datasets of pedestrian motion recorded with consumer-grade IMUs. The TLIO dataset~\cite{liu2020tlio} provides the main quantitative evaluation, while a subset of OxIOD~\cite{chen2020deep} provides a supplementary case study on noisy, uncalibrated smartphone IMU data.
\begin{itemize}
    \item TLIO dataset: We use 205 sequences containing varied human motion recorded with a Bosch BMI055 IMU. The pre-calibrated measurements are downsampled from 1000 to 200~Hz. Fifteen trajectories are held out for testing; the remaining sequences are used for training and validation.
    \item OxIOD subset: We use eight slow-walking sequences recorded with smartphone IMUs at 100~Hz. Models are trained separately on this subset using a 5/2/1 training/validation/test split, as described in Section~\ref{sec:oxiod_case_study}.
\end{itemize}


\subsubsection{Baselines and Models}
We compare two baseline methods representing learning-based IMU integration and filtering with an independently trained neural measurement model:

\begin{itemize}
    \item RoNIN-IncNet: we construct a kinematic dead-reckoning baseline adapted from RoNIN~\cite{herath2020ronin}, denoted as RoNIN-IncNet. Whereas standard RoNIN models regress step-wise velocities for spatial integration, RoNIN-IncNet directly concatenates the 3D position increments predicted by our IncNet model. In this configuration, global orientation is propagated via open-loop gyroscope integration.
    \item TLIO: We use its official open-source implementation~\cite{liu2020tlio}. TLIO fuses learned displacement measurements and covariances within an ESKF architecture, with the neural measurement model trained against local reference targets.
\end{itemize}

We then evaluate four variants of our IncNet--ESKF configurations to examine the effect of filter-aware refinement and the roles of its two output heads:
\begin{itemize}
    \item Phase 1 Only: The IncNet model trained solely on phase 1 losses and then directly deployed in the ESKF.
    \item Tune Inc: Phase~2 refines only the displacement head while keeping the covariance head frozen.
    \item Tune Cov: Phase~2 refines only the covariance head while keeping the displacement head frozen.
    \item DFIO: The complete method jointly refines the displacement and covariance heads in Phase~2.
\end{itemize}
TLIO and Phase 1 Only therefore represent independent measurement learning followed by filtering, whereas Tune Cov, Tune Inc, and DFIO use filter-output supervision during refinement. All four IncNet--ESKF configurations use the same network architecture and filter formulation; their comparison examines the added refinement stage and the effects of trainable heads.

\subsubsection{Metrics}
\label{sec:metrics}
Let $\mathbf p_k$, $\mathbf C_k$, and $\psi_k$ denote the reference position, body-to-world rotation, and yaw at evaluation sample $k$, and let hats denote their estimates. Each trajectory contains $N$ evaluation samples. For relative metrics, $L$ is the integer sample offset corresponding to a fixed time window (e.g., $1\text{ s}$), yielding $M = N - L$ total overlapping evaluation windows. Define $\Delta\mathbf p_{k,L}=\mathbf p_{k+L}-\mathbf p_k$ and $\Delta\psi_{k,L}=\psi_{k+L}-\psi_k$, with analogous definitions for estimated quantities. We report the following metrics:
\begin{itemize}
    \item Absolute translation error (ATE, m): The mean Euclidean position error over a trajectory:
    \begin{equation}
    \mathrm{ATE}=\frac1N\sum_{k=1}^{N}
    \|\hat{\mathbf p}_k-\mathbf p_k\|_2.
    \end{equation}

    \item Relative translation error (RTE-1s, m): The root-mean-square error (RMSE) of the relative translational displacements over fixed-duration ($1\text{ s}$) windows, projected into the start-of-window local body frame:
    \begin{equation}
    \mathrm{RTE}_{1\mathrm s}
    =\sqrt{\frac1M\sum_{k=1}^{M}
    \left\|\hat{\mathbf C}_k^\top\Delta\hat{\mathbf p}_{k,L}
    -\mathbf C_k^\top\Delta\mathbf p_{k,L}\right\|_2^2}.
    \end{equation}

    \item Translation drift rate (TDR, \%): The final position error normalized by reference path length $D=\sum_{k=1}^{N-1}\|\mathbf p_{k+1}-\mathbf p_k\|_2$:
    \begin{equation}
    \mathrm{TDR}=\frac{\|\hat{\mathbf p}_N-\mathbf p_N\|_2}{D}\times100\%.
    \end{equation}

    \item Absolute yaw error (AYE, deg): The RMSE of yaw errors:
    \begin{equation}
    \mathrm{AYE}=\sqrt{\frac1N\sum_{k=1}^{N}
    (\hat\psi_k-\psi_k)^2}.
    \end{equation}

    \item Relative yaw error (RYE-1s, deg): The RMSE of the relative yaw drift over fixed-duration ($1\text{ s}$) windows:
    \begin{equation}
    \mathrm{RYE}_{1\mathrm s}
    =\sqrt{\frac1M\sum_{k=1}^{M}
    (\Delta\hat\psi_{k,L}-\Delta\psi_{k,L})^2}.
    \end{equation}

    \item Yaw drift rate (YDR, deg/h): The final absolute yaw error normalized by trajectory duration $T_h$ in hours:
    \begin{equation}
    \mathrm{YDR}=\frac{|\hat\psi_N-\psi_N|}{T_h}.
    \end{equation}
\end{itemize}

\subsection{Model Performance}
\label{sec:system_comparison}
Table~\ref{tab:tlio_metrics} summarizes the baseline and model-variant results on the TLIO test set, where all entries were recomputed using the definitions in section~\ref{sec:metrics}, with equal weight assigned to each test trajectory in the reported means and medians. The grouped comparison examines how filtering, the pretrained measurement model, and filter-aware head refinement contribute to the performance of the complete method.

\subsubsection{Open-Loop Integration versus Filtering}
\label{sec:filtering_comparison}
All five filtered configurations yield lower mean and median errors than RoNIN-IncNet across the six reported metrics. Even without filter-aware refinement, Phase 1 Only achieves a mean ATE of 1.840~m, compared with 2.657~m for open-loop integration, and a mean YDR of 46.757 compared with 295.233~deg/h. Whereas the open-loop baseline directly accumulates predicted increments and gyroscope rotations, the filter uses displacement residuals and their predicted uncertainty to correct the propagated state.  These results support the benefit of combining learned displacement measurements with recursive IMU-based estimation.

\subsubsection{Independently Trained Measurement Models}
\label{sec:network_comparison}
Having established the benefit of filtering, we compare TLIO with Phase 1 Only. Both train the neural measurement model independently of the filter, and then fuse the displacement measurement and uncertainty within the ESKF. For the network design, IncNet extends TLIO's temporal residual network with temporal self-attention, enabling the measurement model to aggregate information across the input window.

Compared to TLIO, Phase 1 Only reduces mean RYE-1s from 0.893 to 0.231~deg and mean YDR from 197.434 to 46.757~deg/h, corresponding to reductions of 74.1\% and 76.3\%, respectively. Mean ATE and RTE-1s also decrease, from 1.957 to 1.840~m and from 0.140 to 0.102~m. The results show that the independently trained IncNet--ESKF combination already provides improved mean translation and heading accuracy, making it the starting point for DFIO.

\begin{table*}[t]
\TBL{\caption{Results on the 15 held-out TLIO trajectories. Each metric cell reports mean / median values. The lowest mean and median across all six methods are bold.}\label{tab:tlio_metrics}
\label{tab:ablation_study}}
{\begin{fntable}\centering
\footnotesize
\setlength{\tabcolsep}{3pt}
\renewcommand{\arraystretch}{1.3}
\arrayrulecolor{black}
\resizebox{\linewidth}{!}{%
\begin{tabular}{@{}llcccccc@{}}
\hline
Method type & Model & \shortstack{ATE{[m]}} & \shortstack{RTE-1s{[m]}} & \shortstack{TDR{[\%]}} & \shortstack{AYE{[deg]}} & \shortstack{RYE-1s{[deg]}} & \shortstack{YDR{[deg/h]}} \\
\hline
Open-loop & RoNIN-IncNet & 2.657 / 2.197 & 0.244 / 0.218 & 2.852 / 2.985 & 12.903 / 10.983 & 1.723 / 1.333 & 295.233 / 218.982 \\
\hline
 & TLIO & 1.957 / \textbf{1.610} & 0.140 / 0.136 & 1.863 / 1.401 & 2.353 / 1.806 & 0.893 / 0.494 & 197.434 / 73.172 \\
\multirow{-2}{*}{\shortstack[l]{Independent\\learning}} & Phase 1 Only & 1.840 / 1.856 & 0.102 / 0.102 & 1.760 / 2.019 & 2.120 / 1.920 & 0.231 / 0.237 & 46.757 / 45.102 \\
\hline
 & Tune Cov & 1.897 / 1.758 & 0.085 / 0.086 & \textbf{1.479} / \textbf{1.124} & 2.358 / 1.865 & 0.230 / 0.235 & 52.922 / 52.309 \\
 & Tune Inc & 1.795 / 1.631 & 0.085 / 0.085 & 1.673 / 1.540 & \textbf{1.998} / 1.484 & \textbf{0.229} / \textbf{0.231} & 46.214 / 37.220 \\
\multirow{-3}{*}{\shortstack[l]{Filter-aware\\learning}} & DFIO & \textbf{1.699} / 1.645 & \textbf{0.084} / \textbf{0.082} & 1.656 / 1.700 & 2.007 / \textbf{1.370} & 0.231 / 0.233 & \textbf{43.598} / \textbf{37.015} \\
\hline
\end{tabular}%
}
\end{fntable}}
\end{table*}

\subsubsection{Filter-Aware Learning}
\label{sec:phase1_comparison}
The next comparison examines whether refinement using filter-output supervision improves upon independent measurement learning. As shown in Table~\ref{tab:ablation_study}, all three filter-aware variants reduce mean RTE-1s relative to both Phase 1 Only and TLIO: Tune Cov and Tune Inc achieve 0.085~m, while DFIO achieves 0.084~m, compared with 0.102 and 0.140~m for the two independently trained systems. Thus, the refinement stage improves short-interval translation accuracy whether it adjusts the measurement values, their weighting, or both.

For the complete DFIO model, these local gains are accompanied by lower trajectory-wide position errors. Relative to Phase 1 Only, mean ATE decreases by 7.7\%, from 1.840 to 1.699~m, and median ATE decreases from 1.856 to 1.645~m. Mean YDR also decreases by 6.8\%, from 46.757 to 43.598~deg/h, while mean RYE-1s remains 0.231~deg at the reported precision. Filter-aware refinement therefore adds translation-accuracy gains while retaining the heading performance of the pretrained system. Relative to TLIO, the complete method reduces mean ATE by 13.2\% and mean RTE-1s by 39.6\%. The large heading advantage over TLIO is already present after Phase~1; the incremental benefit of Phase~2 is most obvious in translation accuracy.

Table~\ref{tab:paired_results} complements Table~\ref{tab:tlio_metrics} by showing the distribution of improvements across test trajectories. DFIO reduces RTE-1s on all 15 trajectories relative to TLIO and on 14 trajectories relative to Phase~1 Only. The mean paired ATE differences also favor DFIO, at $-0.258$~m and $-0.141$~m, respectively, with lower ATE on 6 and 8 trajectories. Although the 95\% bootstrap intervals for the mean ATE differences span zero, the results demonstrate broad RTE-1s improvements together with lower average position error on the evaluated test set.

\begin{table*}[t]
\centering
\caption{Paired translation-error comparisons
on the 15 TLIO test trajectories. $\Delta$ denotes DFIO minus
the baseline; negative values favor DFIO. The 95\% confidence
intervals are percentile bootstrap intervals for the mean paired
difference, obtained from 100,000 resamples of complete test trajectories}
\label{tab:paired_results}
\small
\setlength{\tabcolsep}{5pt}
\begin{tabular}{llccc}
\hline
Baseline & Metric & Improved / worsened &
Mean $\Delta$ [m] & 95\% CI [m] \\
\hline
TLIO & ATE & $6/9$ &
$-0.2576$ & $[-0.9231,\;0.2861]$ \\
TLIO & RTE-1s & $15/0$ &
$-0.0554$ & $[-0.0718,\;-0.0395]$ \\
Phase 1 Only & ATE & $8/7$ &
$-0.1409$ & $[-0.3510,\;0.0528]$ \\
Phase 1 Only & RTE-1s & $14/1$ &
$-0.0177$ & $[-0.0225,\;-0.0125]$ \\
\hline
\end{tabular}
\end{table*}

\subsubsection{Effects of Displacement and Covariance Refinement}
\label{sec:head_ablation}
The three filter-aware configurations clarify how the displacement and covariance heads contribute. Tune Cov retains the pretrained displacement head yet reduces mean RTE-1s from 0.102 to 0.085~m and achieves the lowest mean and median TDR, 1.479\% and 1.124\%. This shows that adapting measurement weighting can improve local translation accuracy and final-position drift without refining the displacement head. Its mean ATE, however, increases from 1.840 to 1.897~m, and its mean AYE and YDR also increase. Lower final-position drift therefore does not necessarily imply a more accurate trajectory throughout the sequence.

Tune Inc also achieves a mean RTE-1s of 0.085~m and improves mean ATE to 1.795~m. It gives the lowest median ATE among the IncNet--ESKF variants, 1.631~m, and the lowest mean AYE and mean and median RYE-1s among all methods. These results demonstrate the benefit of adapting the predicted displacement itself, even when the covariance head is held fixed.

Joint refinement in DFIO produces the lowest mean ATE, 1.699~m, compared with 1.897~m for Tune Cov and 1.795~m for Tune Inc. It also achieves the lowest mean and median RTE-1s, median AYE, and mean and median YDR across all six methods. In particular, mean YDR decreases to 43.598~deg/h, compared with 52.922 and 46.214~deg/h for the single-head variants. Overall, DFIO improves mean trajectory accuracy and short-interval translation accuracy while offering the lowest heading drift among the evaluated methods.

\begin{figure}
\centering
\includegraphics[width=1.0\linewidth]{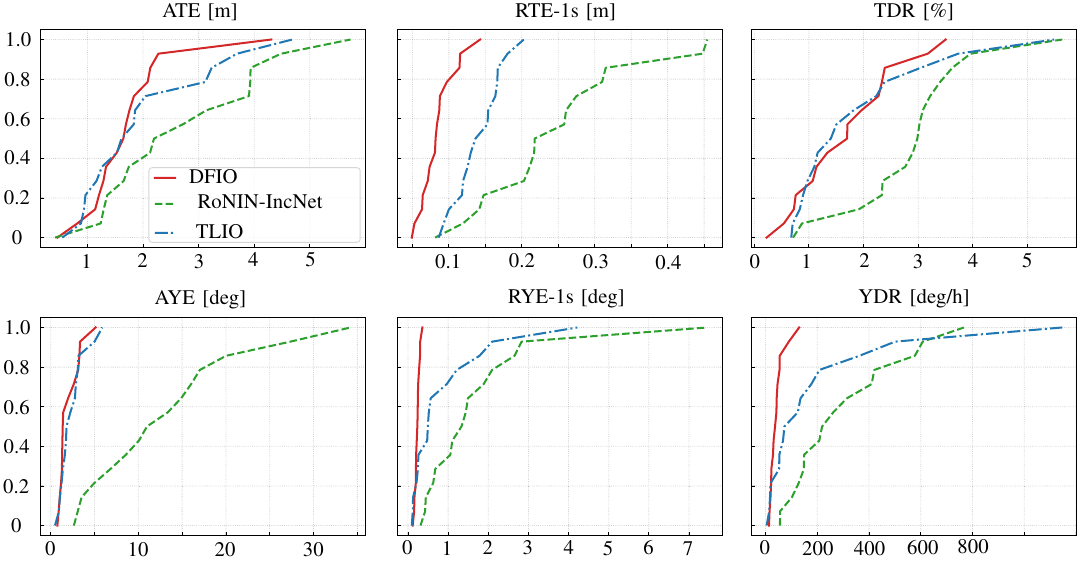}
\caption{Cumulative distributions of error metrics on the TLIO test set for DFIO, TLIO, and RoNIN-IncNet baseline}
\label{fig:tlio_cdf}
\end{figure}

\begin{figure}[thbp!]
\centering
\includegraphics[width=0.9\linewidth]{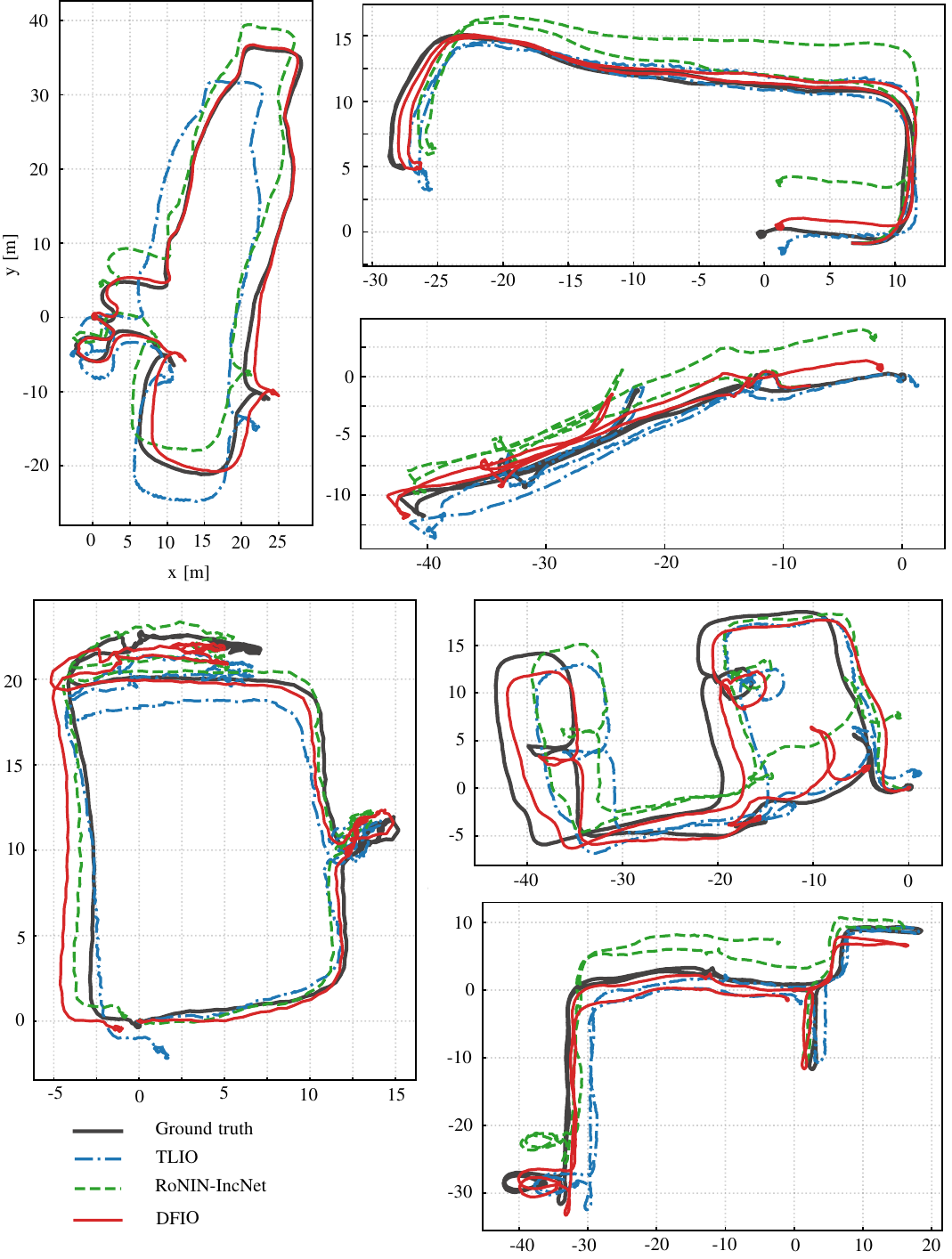}
\caption{Selected planar trajectories from the TLIO test set, comparing reference trajectories, DFIO, TLIO, and RoNIN-IncNet.}
\label{fig:tlio_trajectory}
\end{figure}

Figure~\ref{fig:tlio_cdf} shows the cumulative distribution functions (CDF) of the error metrics for DFIO and the two baselines, and Fig.~\ref{fig:tlio_trajectory} provides selected planar trajectory comparisons.

\subsection{Supplementary OxIOD Case Study}
\label{sec:oxiod_case_study}

To assess the framework's adaptability to significantly different sensor characteristics, we evaluated DFIO on the OxIOD dataset. This data, sourced from uncalibrated, low-cost smartphone IMUs, exhibits high noise levels and bias instability. Both the DFIO and the TLIO models were trained from scratch using the same random split of 5, 2, and 1 sequences for training, validation, and testing, respectively, allowing both networks to learn the specific motion dynamics and noise profiles inherent to the smartphone hardware. Due to the lack of prior calibration data in this dataset, the initial gyroscope and accelerometer biases for the ESKF were initialized to zero.
	
\begin{figure}
\centering
\includegraphics[width=1.0\linewidth]{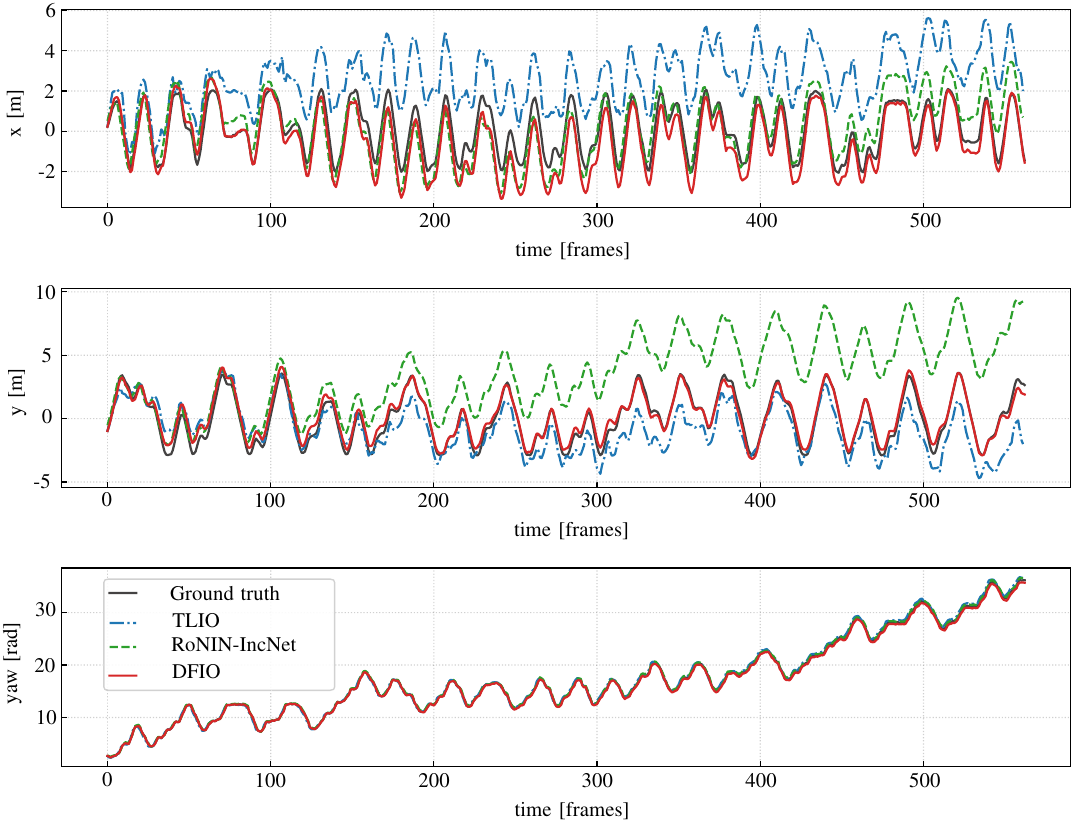}
\caption{Position components and yaw estimation on the held-out OxIOD sequence.}
\label{fig:oxiod_trajectory}
\end{figure}

Figure~\ref{fig:oxiod_trajectory} compares the trajectory components estimated by DFIO, TLIO, and RoNIN-IncNet against the ground truth. While all evaluated models manage to track the general trend of the yaw angle reasonably well throughout the sequence, the baseline methods suffer from significant translational drift. On the contrary, DFIO effectively mitigates this drift, maintaining tight alignment with the ground-truth. 

\section{Conclusions}
This paper presented DFIO, a filter-aware learning framework that combines learned three-dimensional displacement measurements and diagonal covariances with an analytical ESKF. Its two-stage training procedure first learns the measurement model through displacement supervision, then jointly refines the displacement and covariance heads using posterior-position feedback through filter updates. This formulation connects measurement learning to the resulting filtered estimates while retaining analytical IMU propagation.

On 15 held-out TLIO trajectories, DFIO achieved a mean ATE of 1.699~m and a mean RTE-1s of 0.084~m. These correspond to reductions of 13.2\% and 39.6\% relative to the TLIO baseline, and 7.7\% and 17.4\% relative to the independently trained IncNet--ESKF. The refinement primarily improved translation accuracy while retaining the heading accuracy of the pretrained system. Comparisons with the single-head variants showed different trade-offs across the reported metrics, with joint refinement yielding the lowest mean position and relative translation errors among the four IncNet--ESKF model variants. The separately trained OxIOD case study provided a supplementary illustration on a smartphone walking sequence to show DFIO's effectiveness in noisy and uncalibrated IMU data.

The present evaluation concerns pedestrian motion. Future work will investigate a broader range of robotic motions and sensor conditions, together with refinement over longer filter sequences.
	
 \begin{con}
\ctitle{Author Contributions}
Yaru Gu: Conceptualization, methodology, software, formal analysis,
investigation, validation, data curation, visualization,
writing---original draft, and writing---review \& editing. Ting Zou: Conceptualization, supervision, project administration,
funding acquisition, and writing---review \& editing.
 \ctitle{Financial Support}
 This research was supported by the Natural Sciences and Engineering Research Council of Canada (NSERC) through Discovery Grants RGPIN-2019-04131 and RGPIN-2026-04492 awarded to the second author.
 \ctitle{Conflicts of Interest}
The authors declare no conflicts of interest exist.
\ctitle{Data Availability Statement}
The TLIO and OxIOD datasets analysed in this study are publicly available at \url{https://github.com/CathIAS/TLIO} and \url{https://deepio.cs.ox.ac.uk/}, respectively. The DFIO source code will be made publicly available.
 \ctitle{Ethical Approval}
Not applicable.
 \end{con}

\bibliographystyle{robotica-numeric}
	\bibliography{references}

\end{document}